\documentclass[runningheads]{llncs}

\usepackage{verbatim}
\usepackage{multirow}
\usepackage{bbm}
\usepackage{enumitem}
\let\oldput\put
\def\put(#1,#2)#3{%
  \oldput(#1,#2){\sffamily #3}%
}
\usepackage{eccv}
\usepackage{eccvabbrv}

\usepackage{graphicx}
\usepackage{tabularx}
\usepackage{booktabs}
\usepackage{overpic}
\usepackage[accsupp]{axessibility}  

\usepackage{hyperref}

\usepackage{orcidlink}

\begin{document}

\title{ExelMap: Explainable Element-based HD-Map Change Detection and Update\thanks{This work has been accepted for publication at ECCV 2024 2nd Workshop on Vision-Centric Autonomous Driving.} }
\titlerunning{Explainable Element-based Change Detection}

\author{Lena Wild\inst{1,2}\orcidlink{0009-0002-4826-039X} \and Ludvig Ericson\inst{1}\orcidlink{0000-0001-8640-1056} \and
Rafael Valencia\inst{2}\orcidlink{0000-0003-0180-3684}  \and
Patric Jensfelt\inst{1}\orcidlink{0000-0002-1170-7162}}

\authorrunning{L.~Wild et al.}
\institute{Division of Robotics, Perception and Learning, KTH Royal Institute of Technology, Stockholm, Sweden\\
\email{\{lwild, ludv, patric\}@kth.se} \and
Scania CV
AB, Södertälje, Sweden \\
\email{\{lena.wild, rafael.valencia.carreno\}@scania.com}  \vspace{-0.3cm}}

\maketitle
\begin{abstract}
\vspace{-0.25cm}
  Acquisition and maintenance are central problems in deploying high-definition (HD) maps for autonomous driving, with two lines of research prevalent in current literature: Online HD~map generation and HD~map change detection. However, the generated map's quality is currently insufficient for safe deployment, and many change detection approaches fail to precisely localize and extract the changed map elements, hence lacking explainability and hindering a potential fleet-based cooperative HD~map update. 
  In this paper, we propose the novel task of explainable element-based HD~map change detection and update. In extending recent approaches that use online mapping techniques informed with an outdated map prior for HD~map updating, we present ExelMap, an explainable element-based map updating strategy that specifically identifies changed map elements. In this context, we discuss how currently used metrics fail to capture change detection performance, while allowing for unfair comparison between prior-less and prior-informed map generation methods. Finally, we present an experimental study on real-world changes related to pedestrian crossings of the Argoverse 2 Map Change Dataset. To the best of our knowledge, this is the first comprehensive problem investigation of real-world end-to-end element-based HD~map change detection and update, and ExelMap the first proposed solution. 
  \vspace{-0.3cm}
\end{abstract}
\section{Introduction}
\label{sec:intro}

An accurate understanding of the road constitutes the basis for an autonomous vehicle's (AV) downstream applications. A critical component of this technology is High-Definition (HD) maps, which provide a detailed representation of the road environment, facilitating precise navigation beyond solely perception-based reasoning~\cite{hdmaps}. However, the dynamic nature of our world poses a significant challenge: HD~maps must be continuously updated to reflect changes in the environment, a process known as HD~map change detection~\cite{tbv}. \par
Over the last years, high annotation costs and challenges related to the maintainability of maps introduced a shift in HD~map related research, with recent works focusing on sensor-based HD~map \textit{generation} rather than change detection. Online mapping methods aim at extracting topological structure, lane geometry, type and direction, or pedestrian crossings from 360$^{\circ}$-camera images \cite{li2021hdmapnet, vectormapnet, Yuan_2024_streammapnet, maptrv2, MapTR,li2023lanesegnet}. Although promising improvements have been achieved, results are still far from being applicable to real-world driving tasks, as both accuracy and generalizability are not yet sufficient~\cite{loca}. \par
In an attempt to extend the standard map generation pipeline for a  performance boost, recent works explore the integration of, \eg, Standard Definition (SD) Maps~\cite{smerf} or outdated HD~maps~\cite{mindthemap} as a second input besides sensor data. Hence, the network's task is shifted from prior-less generation to fusion of current sensor data and encoded prior map data. Interestingly, if the encoded prior is an outdated HD~map, these works operate at the intersection of online map generation and change detection -- or, as \cite{bateman2024exploringrealworldmap} defines it: HD~map update.\par
Although promising results have been reported for HD~map update, important problems remain unsolved. Firstly, as~\cite{tbv} showed, the low frequency of HD~map elements becoming outdated is a major challenge for both training and evaluation. As real world changes are rare and public datasets do not provide a significant set of stale and up-to-date map pairs, most works resort to synthetic prior noise models~\cite{tbv, mindthemap, bateman2024exploringrealworldmap}. Despite the hope that sufficiently diverse perturbations of the map could bridge the performance gap between training on synthetic data and evaluating the model in the real world -- the so-called sim2real gap --, a significant drop in performance is observed~\cite{bateman2024exploringrealworldmap}. 
\par
A second problem is related to the evaluation of \text{prior-informed} map updating pipelines, and stems from the historic evolution of this line of research. \textit{Prior-informed} map updating works such as~\cite{mindthemap} or \cite{bateman2024exploringrealworldmap} use mAP on all predicted elements, in tradition of \textit{prior-less} map generation frameworks like MapTR~\cite{MapTR}. Not only is this comparison unfair considering that a large part of ground truth map elements are likely to be already present in the encoded stale HD~map, but it also fails to capture the actual change detection or map updating performance by looking at all map elements simultaneously.
\par
Finally, a third problem is present in the only public dataset for change detection: The Argoverse 2 Map Change Dataset~\cite{tbv} -- also known as Trust, but Verify (TbV) --, proposes to determine whether the sensor data is in perfect agreement with the prior map, irrespective of what the potential changes are. Although this metric captures change detection performance, it lacks explainability of which parts of the map have changed. \par
In this work, we show how re-thinking HD~map update as element-based HD~map change detection contributes to tackling these challenges. Our main contributions are 
\vspace{-0.1cm}
\begin{itemize}
    \item to present the novel task of explainable, map element-based change detection and update, as an answer to current challenges,
    \item to propose ExelMap, the first end-to-end explainable element-based HD~map change detection architecture, that includes map updating and
    \item to demonstrate the shortcomings of current evaluation strategies for both change detection and map update while discussing properties of a comprehensive metric.
\end{itemize}
\section{Related work}\label{sec:related}
\subsection{HD-Map Change Detection}\label{sec:change}
Keeping the HD~map up-to-date is a clear requirement for L4 autonomy, as many downstream tasks such as planning and situational awareness consume the map as a fully accurate road representation \cite{hdmaps}. Hence, research efforts have been concentrated on detecting changes in online sensor data compared to a potentially outdated, prior HD~map, as dedicated mapping vehicles cannot update the global HD~map with a high-enough frequency~\cite{tbv, pannen}.  Although such a system would still require the prior acquisition of the HD~map, online change handling would facilitate its maintenance, and prevent detrimental consequences of rare, yet safety critical changes in the road environment. \par
Despite the fact that the field of change detection is an active research direction, it is highly heterogeneous, a consequence of different map definitions and use cases~\cite{plachetka}. Works in HD~map  change detection address different aspects but a unified change detection approach has not yet been presented~\cite{hdmaps, plachetka}. The output of different pipelines varies, from per-pixel change probability~\cite{perpx}, detection of specific ``change categories'' like pedestrian crossings\cite{pedcoss} or roundabouts~\cite{roundabout}, to a binary change score for each input frame~\cite{tbv}.\par
 Another aspect that so far has not been at the center of attention in change detection research is explainability. The change detection task formulated in the only dedicated public dataset, Argoverse 2 Map Change Dataset (TbV)~\cite{tbv}, does not capture what exactly is signaled as a change by the network.  This is detrimental for the safety-critical task of HD~map verification, where human-in-the-loop checks are assumed to be necessary to not constantly degrade the map. Furthermore, identification and localization of changed elements is vital for a potential fleet-based map maintenance effort, to reduce computational load and facilitate information sharing. 
\subsection{HD-Map Generation}\label{sec:maps}
As the acquisition of the HD~maps itself is a significant bottleneck for the scalability of AVs, a different line of research concentrates on learning the local HD~map from sensor data, to avoid annotation costs and mapping downtime. This trend has gained momentum through advances in the development of expressive bird's-eye-view (BEV) feature backbones, that provide a single, compact image feature representation in top down view~\cite{li2022bevformer, Yang2022BEVFormerVA, liu2022bevfusion}.  With earlier works focusing on semantic map segmentation in bird's eye view~\cite{zhou2022crossviewtransformersrealtimemapview}, difficulties arose for complex road topologies, overlapping elements, and their relationships.\par
An early attempt to generate a more compact, \ie, vectorized, local map representation is HDMapNet~\cite{li2021hdmapnet}. Here, an elaborate post-processing consumes the results from semantic segmentation, instance embedding and direction prediction to produce vectorized map elements, yet fails to account for complex relationships between elements. As an answer to this problem, VectorMapNet~\cite{vectormapnet} proposes to directly represent each map element as a sequence of points through a BEV-based decoding scheme, with sequential class-wise key point location extraction. MapTR~\cite{MapTR} and its successor MapTR-v2~\cite{maptrv2} improve inference speed and reduce modeling ambiguity by introducing a unified permutation-equivalent approach to map element decoding.
In extending the information content of the generated HD~map, LaneSegNet~\cite{li2023lanesegnet} combines the detection of map element geometries with the perception of topology relationships, while providing options for integrating semantic lane information, such as lane type. With most prior work relying on single-frame sensor input, StreamMapNet~\cite{Yuan_2024_streammapnet} proposes inter-frame temporal fusion, to increase stability and improve occlusion handling.\par
Although considerable progress has been achieved on the HD~map generation task,
 it can be questioned to what extent the results are applicable to real-world AVs. With map element artifacts being omnipresent in the output, incomplete occlusion handling, and a limited map size of only approximately 100x50 meters around the ego-vehicle, the predicted map fails to meet the requirements of an HD~map prior~\cite{hdmaps}.  As an additional caveat, recent works question the training and evaluation practices on the two commonly used public datasets -- Argoverse 2~\cite{Argoverse2} and nuScenes~\cite{nuscenes} -- , revealing inflated performance of state-of-the-art methods, and an expected performance drop of sometimes more than 45~percentage points in mAP~\cite{Yuan_2024_streammapnet, loca}. 
\subsection{Prior-aided HD-Map Update}\label{sec:novel}
Traces of a potential novel approach at the intersection of change detection and generation of the local map
started to emerge recently. In response to the main map generation problems of low long-range performance and occlusions, integration of priors in the standard sensor-only pipeline has been identified as a viable solution. Such priors are constituted by, \eg, maps of lower accuracy -- so-called standard-definition (SD) maps~\cite{smerf, jiang2024pmapnetfarseeingmapgenerator}--, and stale HD~maps~\cite{mindthemap, bateman2024exploringrealworldmap}. \par
Most related to our work are \cite{bateman2024exploringrealworldmap} and \cite{mindthemap}, which explore the incorporation of stale HD~maps in the HD~map generation pipeline: Rather than maintaining an up-to-date global HD~map, these models reconstruct an updated local representation of the road from both sensor data and the stale map prior. 
Although parallels to the change detection task are manifest, the notion of change does not appear in these settings, as one would need to match the updated map to the prior in post-processing to retrieve changed elements. While such approaches are arguably more streamlined than change detection with subsequent map update, the "implicit" change detection deteriorates verification abilities and control of the network output, as potential passed-through information from the stale HD~map is mixed with newly generated features. \par
The problem of verification is also present in current evaluation strategies, as most works in the field continue using metrics applied in the prior-less map generation task described in~\cref{sec:maps} \cite{bateman2024exploringrealworldmap, mindthemap}. With the frequency of real-world map changes being low, comparison of such \textit{prior-based} map generation techniques with traditional \textit{prior-less sensor-only} approaches masks map updating performance, as with the stale map a large portion of ground truth could be present already in the input.\par
Finally, the last problem concerns the availability and quality of suitable data. Here, it seems that recent works in prior-based map generation face what has been a long known problem in the change detection community: the absence of stale map and up-to-date map pairs\cite{hdmaps, tbv}. As a workaround,~\cite{mindthemap} and~\cite{bateman2024exploringrealworldmap} resort to synthetic ground truth map modification through, \eg, discrete modifications of feature dropout or duplication, or continuous modifications of applying noise or warps. However, this results in a considerable sim2real gap when evaluating on real changes~\cite{bateman2024exploringrealworldmap}. 
\par
\vspace{0.2cm}
In this paper, we argue that some of these challenges can be either solved or avoided in first place in the novel task of explainable change detection and update for HD~map maintenance. In fact, as the general decoding paradigm of map prediction approaches operates on an element-wise basis, assuring explainability through element-wise change assessment is intrinsically possible, while map update follows seamlessly from the pipeline's output. Finally, although the dataset and evaluation problem remain to be solved decisively, an element-based approach combined with an adequate change detection metric could lead to a better understanding of the underlying challenges.

\section{Methodology}
Our approach is based on LaneSegNet~\cite{li2023lanesegnet}, but it can be flexibly adapted to other state-of-the-art map generation methods. We choose LaneSegNet because the generated map includes semantic attributes like lane type by default, and is easily extendable to prediction of map elements in forward view, such as traffic lights and signs. Furthermore, the network operates on the basis of so-called lane segments, which are represented as a vectorized centerline and corresponding lane boundaries, \(\mathcal{V} = \{\mathcal{V}_\text{center}, \mathcal{V}_\text{left}, \mathcal{V}_\text{right}\}\). The class of the respective map element -- lane or pedestrian crossing -- is then determined by the output of a classification head \cite{li2023lanesegnet}. This unified representation is beneficial for designing a change detection pipeline that can flexibly detect changes in both lanes and pedestrian crossings simultaneously. \cref{fig:example} shows the architecture of our method ExelMap. In the following subsections, we will detail where our method extends the LaneSegNet backbone, to facilitate reproduction of our results. 
\subsection{Stale HD-Map Encoder}
In order to integrate the stale HD~map in the LaneSegNet backbone, we leverage the lightweight SD map tokenizer of~\cite{smerf} and adapt it to our HD~map requirements. Our stale HD~map contains geometric information of the local lane segments, their class (pedestrian crossing or lane) and their semantic attributes (right and left lane border type, \ie, dashed, solid or non-visible). Additionally, lane segment connectivity is stored in the local road graph, but this is particularly costly to annotate and keep up-to-date. Hence, we leave the prediction of the local graph structure to the generation part of the pipeline, to alleviate requirements of the HD~map prior.\par
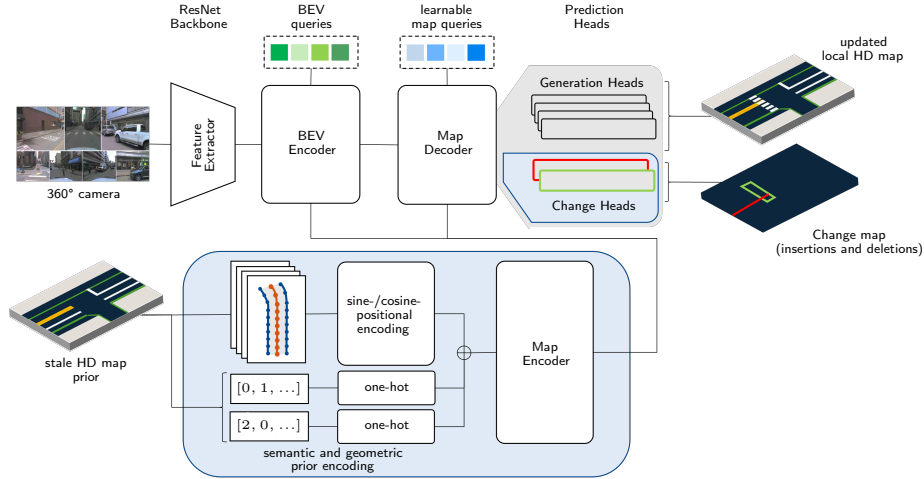
\begin{figure}[tb]
  \centering
  \begin{overpic}[width=\textwidth, grid=False]{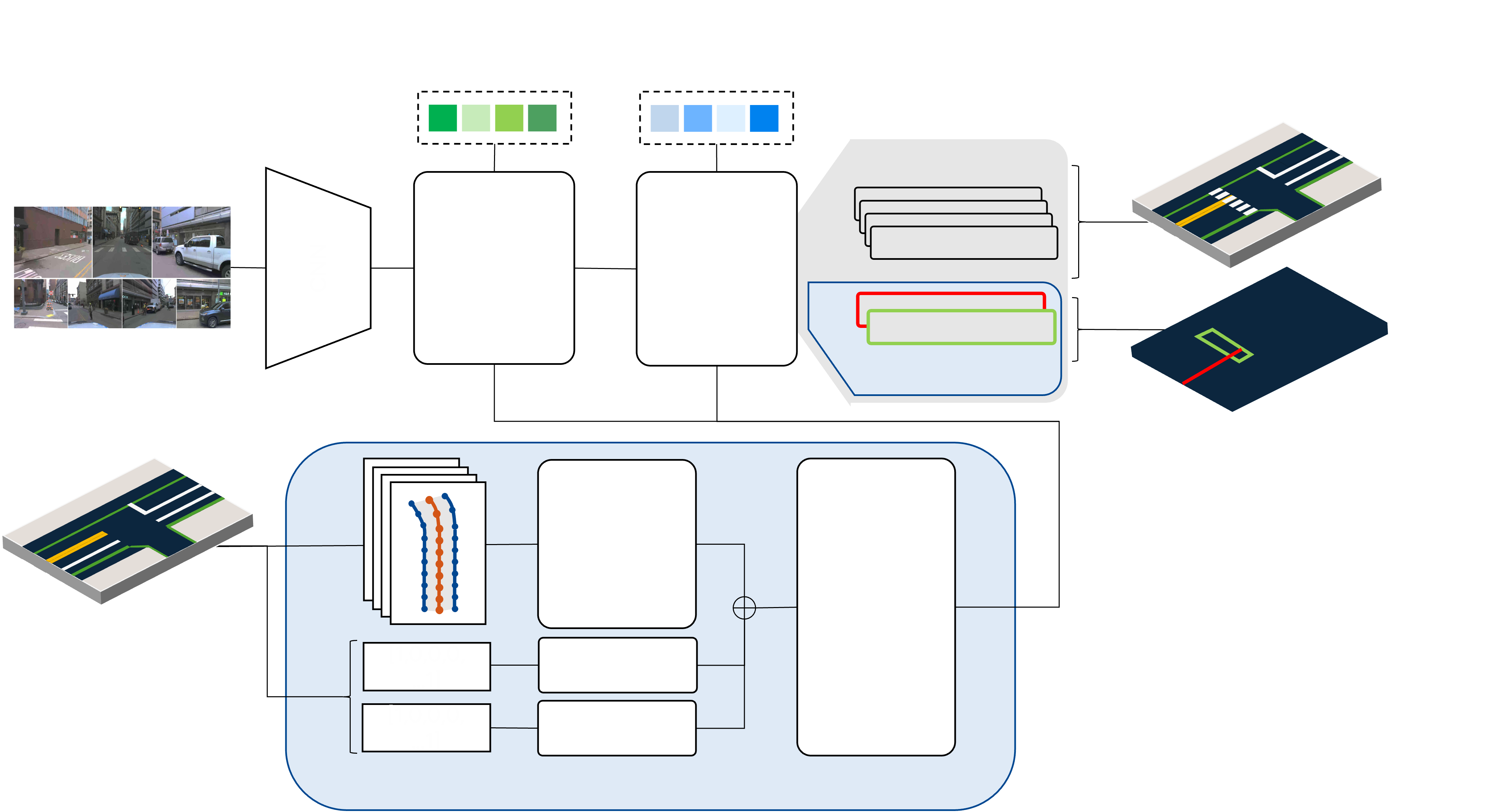}
  \put(4,12){\tiny stale HD~map}
  \put(7,10.5){\tiny prior}
  \put(4.2,30.5){\tiny 360° camera}
  \put(24.8,9.26){\tiny \([0,1,...]\)}
  \put(24.8,5.26){\tiny \([2,0,...]\)}
  \put(27.8,2.26){\tiny semantic and geometric}
  \put(30.8,0.76){\tiny prior encoding}
  \put(38.8,9.46){\tiny one-hot}
  \put(38.8,5.26){\tiny one-hot}
  \put(36.9,18.76){\tiny sine-/cosine-}
  \put(37.9,17.26){\tiny positional}
  \put(38.0,15.76){\tiny encoding}
  \put(57.0,13.76){\tiny Map}
  \put(56.0,12.26){\tiny Encoder}
  \put(20.0,33.76){\rotatebox{90}{\tiny Feature}}
  \put(21.4,33.29){\rotatebox{90}{\tiny Extractor}}
  \put(31.50,36.79){\tiny BEV}
  \put(30.50,35.29){\tiny Encoder}
  \put(31.50,50.5){\tiny BEV}
  \put(30.80,49.0){\tiny queries}
  \put(44.90,50.5){\tiny learnable}
  \put(43.90,49.0){\tiny map queries}
  \put(46.7,36.79){\tiny Map}
  \put(44.40,35.29){ \tiny Decoder}
  \put(60.60,50.5){\tiny Prediction}
  \put(61.60,49.0){\tiny Heads}
  \put(18.60,50.5){\tiny ResNet}
  \put(17.60,49.0){\tiny Backbone}
  \put(59.2,29.10){\tiny Change Heads}
  \put(57.9,42.60){\tiny Generation Heads}
  \put(90.6,47.0){\tiny updated}
  \put(88.6,45.5){\tiny local HD~map}
  \put(87.6,26.0){\tiny Change map}
  \put(83.6,24.5){\tiny (insertions and deletions)}

  \end{overpic}
  
  \caption{Architecture of ExelMap. The HD~map encoder is based on \cite{smerf} with our own extensions. In addition to the standard map generation heads, the insertion and deletion heads detect changes element-wise. The output of the network is twofold: an updated local HD~map, and a second map containing the element-wise change score.}
 
  \label{fig:example}
\end{figure}

 
We start by extracting the geometrical representation of all \(m_{\text{prior}}\) lane segments in the outdated HD~map,  which as defined in \cite{li2023lanesegnet} consist of 10 equally spaced points for left lane, right lane, and centerline, which we denote by \(\mathcal{V}=\{(x_i, y_i)\}^{30}_{i=1}\). Additionally, we extract left and right lane boundary-type, \ie, not visible, dashed or solid, as shown in ~\cref{fig:example}. We choose a sine/cosine positional encoding to adequately capture small curvatures following~\cite{smerf} for the \(m_\text{prior}\) spatial coordinates, and a one-hot encoding scheme of dimension \(k=3\) for right and left lane boundary-type. Subsequently, we concatenate the encoded geometrical coordinates and the one-hot-encoded boundary types to achieve the desired polyline sequence of shape \(m_\text{prior} \times (30\cdot d + 2\cdot k)\), with \(d\) being the positional embedding dimension. Subsequently, the stacked, encoded input is fed into a transformer map encoder, which consists of 6 layers with a self-attention block and a feed-forward network following \cite{smerf}.
\subsection{Encoded Map Prior Integration}
To consume the encoded stale map features in the map prediction pipeline, two lines of thought are prevalent in  literature: (1) to attend to the encoded map in an additional cross-attention step, or (2) to use the encoded map as queries in the final map decoder model. Both approaches have intuitive interpretations: In (1), the stale map is merely a second source of information alongside the sensor data encoded in the BEV features. In (2),  the map prior is used as decoder tokens, which are subsequently refined by attending to the BEV features. \par
The two works most related to ours,~\cite{mindthemap} and \cite{bateman2024exploringrealworldmap}, choose the second option and argue that this allows the model to meaningfully leverage prior information. However, the authors of~\cite{mindthemap} mention that the model sometimes does not account for existing map information, and that the network can fail to identify even fully accurate prior-based queries, if left on its own, which they solve by pre-attributing predictions to corresponding stale map element in an additional matching step. The authors of~\cite{bateman2024exploringrealworldmap} report that for more complex deviations in the prior, the model returns to reproducing the prior only. 
This is consistent with our observations, and we suspect the  behavior to be caused by the strong imbalance between changed and unchanged elements in stale maps, which makes learning a pass-through function tempting yet detrimental for larger map changes. \par
Hence, we opt for strategy (1), and design fully learnable map decoder queries to give the network enough flexibility for incorporating sensor data and stale map equally. To avoid what~\cite{bateman2024exploringrealworldmap} mentions as the main problem in this approach -- that cross-attention
does not provide enough bandwidth for the model to incorporate the prior as strongly as it should -- we propose a novel double cross-attention integration.  One can think of two locations in the pipeline to leverage the prior: in the BEVformer-based~\cite{li2022bevformer, Yang2022BEVFormerVA} transformer encoder, that transforms the sensor feature representation of the ResNet-backbone~\cite{resnet} to bird's eye view, or in the map decoder, that extracts the map feature representation by cross-attending to the BEV features. Contrarily to~\cite{smerf},  we choose to cross-attend after each spatial cross-attention in the BEV encoder \textit{and} in the map decoder.
\subsection{Change Detection Heads}
To adapt the LaneSegNet backbone to explainable element-based change detection, we extend the standard map prediction branches to extract the change status of each element alongside the geometric, semantic and topological aspects from all \(m_\text{pred}\) lane segment queries. 
We choose to add two distinct, independently operating heads with binary output for element-wise deletion and insertion detection.  As loss function, we found the Focal Loss~\cite{focalloss} to be suitable, as we have a strong class imbalance between changed and unchanged map elements, and between frames containing a change or not. Given the lane segment based approach, the class of the changed element -- pedestrian crossing or lane -- is determined independently by the standard classification head. \par
Both change detection heads consist of a series of linear layers with ReLU-activation and dropout, and output a tensor of shape \((m_\text{pred}, 1)\). We choose to separate heads for deletion and insertion detection, as binary classification is less challenging for the network. Surprisingly, we never observe conflicting behavior between the two heads, although they operate in parallel. 
\par
With these additional heads, the network's output is twofold: An updated HD~map representation of the road scene in the tradition of works in HD~map update, and a novel change map, where each predicted element has a change status -- \ie, unchanged, inserted or deleted -- allowing for explainable human-in-the-loop change verification in this safety critical area. \\

\subsection{Datasets and Synthetic Change Generation}
As the authors of the only publicly available HD~map change detection dataset TbV \cite{tbv} point out, there is a substantial gap between the development and training of change detection algorithms, and their evaluation on real-world changes. To fill this gap,  TbV \cite{tbv} is the first to provide more than 200 scenarios with real-world changes, mainly related to road geometry change or change of semantic lane attributes, and pedestrian crossings. For training, they once again propose to synthetically alter ground truth maps. \par
We therefore choose to evaluate our network on the real-world map changes of TbV, and to train on the proposed train split with no real-world map changes. Although our approach is able to detect changes in \textit{all} lane segments, \ie, both with respect to pedestrian crossings and lanes, we choose to focus on pedestrian crossing-related changes only for our present experiments. This is because missing or newly painted pedestrian crossings are a frequent map change, easy to synthetically generate and prominent features to detect and evaluate. Hence, in training, we randomly delete individual pedestrian crossings. For insertions, we manually edit the maps instead of using the automatic toolkit proposed in \cite{tbv}, because we found that the latter cannot provide a suitable variety of realistic map changes. \par
Finally, we pre-process the training data to match the specifications of the most used map generation datasets, \ie, Argoverse 2 \cite{Argoverse2}, nuScenes \cite{nuscenes} and Openlane-V2 \cite{wang2023openlanev2}. We reformulate the criteria of a lane segment start- or endpoint as either a change in connectivity or semantic lane attributes, to have longer lane segments for a facilitated training process compared to the original dataset following \cite{wang2023openlanev2}. Given that we are not interested in longer-range map generation but rather in scanning the road for changes, we reduce the field of view to [\(-\)25m, +25m] along both x and y-axis.
\section{Experiments}
For all configurations of our model, we employ pretrained ResNet-50 \cite{resnet} as an image backbone. From the model output, we compute the one-to-one optimal assignment with the Hungarian algorithm. We train the model for 20 epochs with a batch size of 8 and AdamW optimizer on 8 NVIDIA A10G Tensor Core GPUs. From the annotated validation split, we extract all 66 real-world examples of pedestrian crossing-related changes, which are present in 33 driving sequences, and all four sequences with no change for a total of approximately 3800 frames. Of these examples, 46 are pedestrian crossing deletions, and 20 are insertions. We do not consider shifts of elements, as they are not present in the validation set in the first place. Furthermore, these changes could be easily formulated as deletion with subsequent insertion. 
\begin{figure}[th]
  \centering
  \begin{overpic}[width=1.0\linewidth, grid=False]{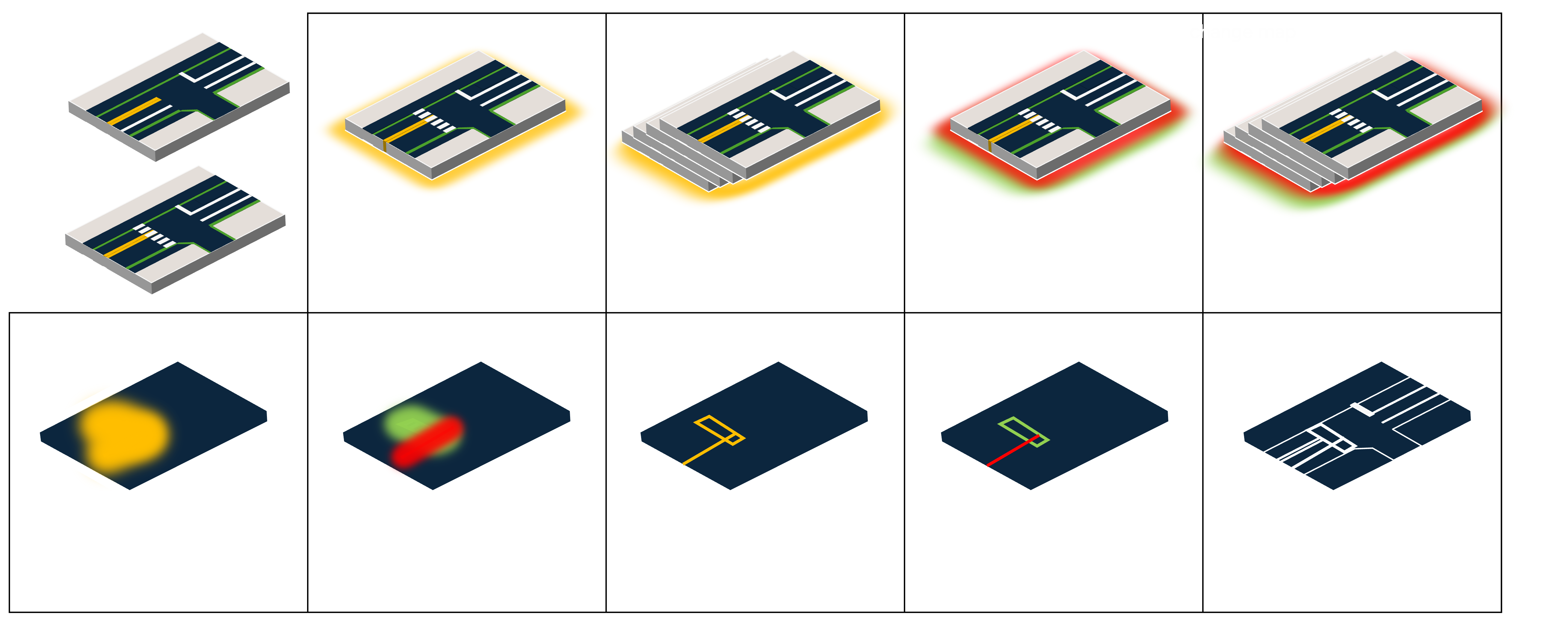}
  \put(4.4, 5){\fontsize{6}{6}\selectfont Type-Agnostic}
  \put(2.5, 3.3){\fontsize{6}{6}\selectfont Change Localization}
  \put(24.6, 5){\fontsize{6}{6}\selectfont Type-Aware}
  \put(21.6, 3.3){\fontsize{6}{6}\selectfont Change Localization}
  \put(43.1, 5){\fontsize{6}{6}\selectfont Type-Agnostic }
  \put(41.5, 3.3){\fontsize{6}{6}\selectfont Element Accuracy}
  \put(63.5, 5){\fontsize{6}{6}\selectfont Type-Aware}
  \put(60.5, 3.3){\fontsize{6}{6}\selectfont Element Accuracy}
  \put(81.3, 5){\fontsize{6}{6}\selectfont Accuracy of }
  \put(80.8, 3.3){\fontsize{6}{6}\selectfont Updated Map}
  
  \put(23.5, 24){\fontsize{6}{6}\selectfont Type-Agnostic }
  \put(21, 22.3){\fontsize{6}{6}\selectfont Change Detection, SF}
  \put(43.3, 24){\fontsize{6}{6}\selectfont Type-Agnostic }
  \put(39.5, 22.3){\fontsize{6}{6}\selectfont Change Detection, MF}
  \put(62.5, 24){\fontsize{6}{6}\selectfont Type-Aware }
  \put(58.6, 22.3){\fontsize{6}{6}\selectfont Change Detection, SF}
  \put(81.5, 24){\fontsize{6}{6}\selectfont Type-Aware }
  \put(77.3, 22.3){\fontsize{6}{6}\selectfont Change Detection, MF}
  
  \put(1.3, 36.8){\fontsize{6}{6}\selectfont Map Prior}
  \put(1.7, 27.8){\fontsize{6}{6}\selectfont GT Map}

  \put(19.8, 37.5){\fontsize{6}{6}\selectfont (a)}
  \put(38.8, 37.5){\fontsize{6}{6}\selectfont (b)}
  \put(57.9, 37.5){\fontsize{6}{6}\selectfont (c)}
  \put(77.0, 37.5){\fontsize{6}{6}\selectfont (d)}

  \put(0.75, 18.3){\fontsize{6}{6}\selectfont (e)}
  \put(19.8, 18.3){\fontsize{6}{6}\selectfont (f)}
  \put(38.8, 18.3){\fontsize{6}{6}\selectfont (g)}
  \put(57.9, 18.3){\fontsize{6}{6}\selectfont (h)}
  \put(77.0, 18.3){\fontsize{6}{6}\selectfont (i)}
  
  \end{overpic}
  
  \caption{Evaluation Strategies for explainable element-based Change Detection. Yellow highlights represent change-type agnostic scores, red highlights indicate lane segment deletions, green highlights indicate pedestrian crossing insertions. We differentiate between single frame (SF) and multi-frame evaluation. A mathematical description of (a)--(i) is provided in~\cref{sec:mathe}.  
  }
  \label{fig:changes}
\end{figure}
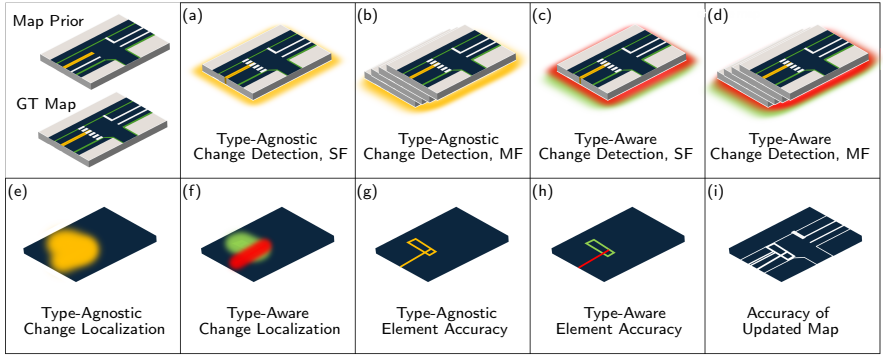
\subsection{Explainable Change Detection Metrics}\label{sec:mathe}
The task we propose tackles novel problems, and hence comparison to prior work is difficult. Furthermore, we cannot simply apply metrics from HD~map generation or HD~map update, due to their shortcomings discussed in~\cref{sec:change,sec:novel}. Instead, we provide a discussion of nine network qualities a new metric for this task should capture, and evaluate our approach accordingly. These properties are in parts inspired by the map generation metric as used in \cite{MapTR} and others, as well as in the metric established in the TbV dataset presentation \cite{tbv}. \par
The first question an adequate metric should answer is how well the network detects that there is a change / no change in the current input, irrespective of what the change is, \ie, change type-agnostic (\cref{fig:changes},(a)). This is similar to  the evaluation of \cite{tbv} for multi-frame (MF), \ie, sequential input, and accounts for precision and recall in this two-class problem. Given that our approach operates on a single-frame (SF) basis, we evaluate the SF type-agnostic change detection accuracy \(\text{Acc}^{+}\) and no-change detection accuracy \(\text{Acc}^{-}\) by adapting \cite{tbv} to 
\begin{equation}\label{eq:1}
\text{Acc}^{+(-)} = \frac{\sum\limits_{i=0}^{\mathcal{S}}\sum\limits_{j=0}^{\mathcal{F}^{i}}  \mathbbm{1}\{ \hat{y}_{ij}^{\text{SF}} = y_{ij}^{\text{SF}}\} \cdot \mathbbm{1}\{ y_{ij}^{\text{SF}} = c \ (\neg c) \}  }{\sum\limits_{i=0}^{\mathcal{S}}\sum\limits_{j=0}^{\mathcal{F}^{i}}  \mathbbm{1}\{y_{ij}^{\text{SF}} = c \ (\neg c) \} },
\end{equation}
with
 \begin{align}\label{eq:2}
  \hat{y}_{ij}^{\text{SF}} &=\begin{cases}
    1, & \text{if $\sum\limits_{k=0}^{m_\text{pred}^{ij}} \mathbbm{1}\left\{\left(\hat{y}_{ijk}^{\text{del}} = c \right) \vee \left(\hat{y}_{ijk}^{\text{ins}} = c\right)\right\} \cdot \mathbbm{1}\left\{\hat{y}_{ijk}^{\text{score}} \geq  \epsilon \right\} $} > 0\\
    0, & \text{otherwise}\end{cases},
\end{align}
\normalsize
where \(\mathcal{S}\) denotes the number of sequences, \(\mathcal{F}^{i}\) the number of frames in the \(i\)-th sequence, and \(m_\text{pred}^{ij}\) the number of elements in the \(j\)-th frame of the \(i\)-th sequence. \(c\) and \(\neg c\) are 1 and 0 respectively, and represent change or no change. \(y_{ij}^{\text{SF}}\) is the ground truth SF type-agnostic change score per frame. The SF prediction \(\hat{y}_{ij}^{\text{SF}}\) is 1 if insertion head \(\hat{y}_{ijk}^{\text{ins}}\) or deletion head \(\hat{y}_{ijk}^{\text{del}}\) signals a change for any of the predicted elements in \(m_\text{pred}^{ij}\), if the predicted element score \(\hat{y}_{ijk}^{\text{score}}\) is above a preset threshold \(\epsilon\). The mean accuracy is then given by \(\text{mAcc} =  (\text{Acc}^{+} + \text{Acc}^{-})/2\).\par
For MF type-agnostic change detection accuracy (\cref{fig:changes},(b)), we define
\begin{equation}
\text{Acc}^{+(-)} = \frac{\sum\limits_{i=0}^{\mathcal{S}} \mathbbm{1}\{ \hat{y}_i^{\text{MF}} = y_i^{\text{MF}}\} \cdot \mathbbm{1}\{ y_i^\text{MF} = c \ (\neg c) \}  }{\sum\limits_{i=0}^{\mathcal{S}} \mathbbm{1}\{y_i^{\text{MF}} = c \ (\neg c) \} },
\end{equation}
\normalsize with 
 \begin{align}
  \hat{y}_i^{\text{MF}} &=\begin{cases}
    1, & \hspace{-0.0cm}\text{if $\sum\limits_{j=0}^{\mathcal{F}^{i}}\sum\limits_{k=0}^{m_\text{pred}^{ij}} \mathbbm{1}\left\{\left(\hat{y}_{ijk}^{\text{del}} = c \right) \vee \left(\hat{y}_{ijk}^{\text{ins}} = c\right)\right\} \cdot \mathbbm{1}\left\{\hat{y}_{ijk}^{\text{score}} \geq  \epsilon \right\} $} > 0\\
    0, & \hspace{-0.0cm} \text{otherwise}\end{cases}.
\end{align}\par
\normalsize
For SF and MF type-aware change detection accuracy, \ie, indicating that there is a change of a certain type in the frame or sequence, we evaluate~\cref{eq:1} change-type specific (\cref{fig:changes},(c)--(d)). We define \(c_t=1\) for a change of a specific type \(t\), where in our case \(t=\{\)pedestrian crossing insertion, pedestrian crossing deletion\(\}\). This leads to the following adaptations for our predictions: 
 \begin{align}\label{eq:3}
  \hat{y}_{ij}^{\text{SF,ins(del)}} &=\begin{cases}
    1, & \text{if $\sum\limits_{k=0}^{m_\text{pred}^{ij}} \mathbbm{1}\left\{ \hat{y}_{ijk}^{\text{ins(del)}} = c_\text{ins(del)}\right\} \cdot \mathbbm{1}\left\{\hat{y}_{ijk}^{\text{score}} \geq  \epsilon \right\} $} > 0\\
    0, & \text{otherwise}\end{cases},
\end{align}
and
 \begin{align}
  \hat{y}_i^{\text{MF,ins(del)}} &=\begin{cases}
    1, & \text{if $\sum\limits_{j=0}^{\mathcal{F}^{i}}\sum\limits_{k=0}^{m_\text{pred}^{ij}} \mathbbm{1}\left\{\hat{y}_{ijk}^{\text{ins(del)}} = c_\text{ins(del)}\right\} \cdot \mathbbm{1}\left\{\hat{y}_{ijk}^{\text{score}} \geq  \epsilon \right\}  $} > 0\\
    0, & \text{otherwise}\end{cases}.
\end{align}\par
\normalsize
To determine how well the detected changes are localized compared to ground truth (\cref{fig:changes},(e)--(f)), we compute the intersection over union (IoU) between predicted changed elements and ground truth changed elements.
For every detected changed element in \(\mathcal{S}_i\) and \(\mathcal{F}_j\), 
\begin{equation}\hat{y}_{ijk}^\text{changed} = (\{\hat{y}_{ijk}^{\text{ins}}=1\} \vee \{\hat{y}_{ijk}^{\text{del}}=1\}),
\end{equation} we consider it localized if 
\begin{align}
\sum\limits_{l=0}^{m_\text{GT}^{ij}} \mathbbm{1}\left\{ \text{IoU}(\hat{y}_{ijk}^\text{changed}, y_{ijl}^\text{changed}) \geq \theta \right\} \cdot \mathbbm{1}\left\{\hat{y}_{ijk}^{\text{score}} \geq  \epsilon \right\}  > 0,
\end{align}
\normalsize
where \(y_{ijl}^\text{changed}\) is a ground truth changed element and \(\theta\) the IoU threshold. The localization accuarcy \(\text{Acc}_\text{loca}\) is determined by dividing the number of localized changed elements by the number of detections. For the class-aware evaluation in \cref{fig:changes},(f), we separate insertions and deletions by defining  \(\hat{y}_{ijk}^\text{changed} = \{\hat{y}_{ijk}^{\text{ins(del)}}=1\}\). In a slightly different evaluation mode, we calculate the accuracy only for frames in which there is at least one changed ground truth element \(y_{ijl}^\text{changed}\), \(\text{Acc}_\text{loca}^c\). The difference between \(\text{Acc}_\text{loca}\) and \(\text{Acc}_\text{loca}^c\) allows for insights into the amount of false positive detections in this localization setting.\par
Finally, to determine the AP of changed elements, both type-agnostic and type-aware (\cref{fig:changes},(g)--(h)), and the AP of the updated map in tradition of HD~map generation and HD~map update research, we use the metric used in the LaneSegNet backbone \cite{li2023lanesegnet}. Note that~\cref{fig:changes},(i) is similar to evaluation practices in \cite{mindthemap, bateman2024exploringrealworldmap}. The distance between elements as a weighted sum of the distances between left and right lane boundaries and centerlines, can be written as
\begin{equation}
    \mathcal{D}(\mathcal{V}, \widetilde{\mathcal{V}}) = 
    \frac{1}{2} \Big[ \texttt{Chamfer}([\mathcal{V}_{\text{left}}, \mathcal{V}_{\text{right}}], 
    [\widetilde{\mathcal{V}}_{\text{left}}, \widetilde{\mathcal{V}}_{\text{right}}]) 
    + \texttt{Fr\'{e}chet}(\mathcal{V}_{\text{ctr}}, \widetilde{\mathcal{V}}_{\text{ctr}} )\Big],
\end{equation}
\normalsize
with prediction $\mathcal{V}$ and the ground truth $\widetilde{\mathcal{V}}$. If the element is a pedestrian crossing, only the non-directional Chamfer Distance is used. 
The final AP score is calculated over the three matching thresholds \(\mathbbm{T} = \{1.0, 2.0, 3.0\}\) meters for lanes and \(\mathbbm{T} = \{0.5, 1.0, 1.5\}\) meter for pedestrian crossings as defined in \cite{li2023lanesegnet}.\par
A comprehensive evaluation of our model under all nine evaluation strategies is provided in~\cref{table2}. For (a)--(d), we evaluate for an element-score threshold of \(\epsilon=\{0.2,0.3,0.4\}\). For (e) and (f), we set \(\epsilon=0.3\) and evaluate for a IoU threshold of \(\theta=\{0.3, 0.5, 0.8\}\). 
\begin{table}[htbp]
\caption{Evaluation of ExelMap according to evaluation strategies (a)--(i) depicted in~\cref{fig:changes} and detailed in \cref{sec:mathe}. \(\epsilon\) is the minimum element-wise score for element acceptance in the output, \(\theta\) is the threshold for IoU. SF denotes single frame evaluation, MF multi frame (\(^*\)as our approach is inherently SF, we evaluate a pseudo-MF approach by looking at the respective sequence of SF predictions without temporal fusion).}
\label{table2}
\centering
\begin{tabular}{@{}c|c|c|c||ccc@{}}
\bottomrule
\toprule
strategy & modality & change class & \(\epsilon\) &\(\text{Acc}^{+}\) & \(\text{Acc}^{-}\)  & mAcc  \\ \midrule

\multirow{ 3}{*}{(a)} &\multirow{ 3}{*}{SF} & \multirow{3}{*}{All Ped.-crossing Changes} & 0.2 & \textbf{0.52} & 0.87 & \textbf{0.70}  \\
 & & & 0.3 & 0.43 & 0.93 & 0.68  \\ 
  & & & 0.4 & 0.37 & \textbf{0.97} & 0.67  \\ 
   \midrule

\multirow{ 3}{*}{(b)} & \multirow{ 3}{*}{\(\text{MF}^*\)} & \multirow{3}{*}{All Ped.-crossing Changes} & 0.2 & 0.94 & 0.00 & 0.47  \\
 & & & 0.3 & 0.94 & 0.25 & 0.60  \\ 
  & & & 0.4 & \textbf{0.94} & \textbf{0.75} & \textbf{0.84}  \\ 
   \midrule

\multirow{ 6}{*}{(c)} & \multirow{ 6}{*}{SF} & \multirow{3}{*}{Ped.-crossing Insertion} & 0.2 & \textbf{0.87} & 0.86 & \textbf{0.86}  \\
 & & & 0.3 & 0.79 & 0.93 & 0.86  \\ 
  & & & 0.4 & 0.68 & \textbf{0.97} & 0.83  \\ 
 \cmidrule(){3-7}
 & & \multirow{3}{*}{Ped.-crossing Deletion} & 0.2 & \textbf{0.12} & \textbf{0.99} & \textbf{0.56}  \\
 & & & 0.3 & 0.11 & 0.99 & 0.55  \\ 
  & & & 0.4 & 0.11 & 0.99 & 0.55 \\ 
   \midrule

\multirow{ 6}{*}{(d)} & \multirow{ 6}{*}{\(\text{MF}^*\)} & \multirow{3}{*}{Ped.-crossing Insertion} & 0.2 & 0.93 & 0.05 & 0.49  \\
 & & & 0.3 & 0.93 & 0.18 & 0.56  \\ 
  & & & 0.4 & \textbf{0.93} & \textbf{0.45} & \textbf{
0.69} \\ 
 \cmidrule(){3-7}
 & & \multirow{3}{*}{Ped.-crossing Deletion} & 0.2 & \textbf{0.47} & \textbf{0.83} & \textbf{0.65}  \\
 & & & 0.3 & 0.42 & 0.83 & 0.63  \\ 
  & & & 0.4 & 0.42 & 0.83 & 0.63  \\ 
\bottomrule
\toprule
strategy & modality & change class & \(\theta\) &  \multicolumn{ 3}{l}{\begin{tabular}{>{\centering}p{0.09\textwidth}>{\centering}p{0.09\textwidth}}
  \(\text{Acc}_\text{loca }\)   & \(\text{Acc}_\text{loca}^{\text{c}}\) 
\end{tabular}}\\ 
\midrule

\multirow{ 3}{*}{(e)} & \multirow{ 3}{*}{\(\text{SF}\)} & \multirow{3}{*}{All Ped.-crossing Changes} & 0.3 & \multicolumn{ 3}{l}{\begin{tabular}{>{\centering}p{0.09\textwidth}>{\centering}p{0.09\textwidth}}
  0.63 \qquad & 0.79 
\end{tabular}}\\
 & & & 0.5 & \multicolumn{ 3}{l}{\begin{tabular}{>{\centering}p{0.09\textwidth}>{\centering}p{0.09\textwidth}}
  0.47  & 0.59 
\end{tabular}}\\
  & & & 0.8 &\multicolumn{ 3}{l}{\begin{tabular}{>{\centering}p{0.09\textwidth}>{\centering}p{0.09\textwidth}}
  0.13  & 0.16 
\end{tabular}}\\
   \midrule

\multirow{ 6}{*}{(f)} & \multirow{ 6}{*}{SF} & \multirow{3}{*}{Ped.-crossing Insertion} & 0.3 & \multicolumn{ 3}{l}{\begin{tabular}{>{\centering}p{0.09\textwidth}>{\centering}p{0.09\textwidth}}
  0.60  & 0.87 
\end{tabular}}\\
 & & & 0.5 & \multicolumn{ 3}{l}{\begin{tabular}{>{\centering}p{0.09\textwidth}>{\centering}p{0.09\textwidth}}
  0.41  & 0.60 
\end{tabular}}\\
  & & & 0.8 & \multicolumn{ 3}{l}{\begin{tabular}{>{\centering}p{0.09\textwidth}>{\centering}p{0.09\textwidth}}
  0.07  & 0.11 
\end{tabular}}\\
 \cmidrule(){3-7}
 & & \multirow{3}{*}{Ped.-crossing Deletion} & 0.3 & \multicolumn{ 3}{l}{\begin{tabular}{>{\centering}p{0.09\textwidth}>{\centering}p{0.09\textwidth}}
  0.90  & 0.99 
\end{tabular}}\\
 & & & 0.5 & \multicolumn{ 3}{l}{\begin{tabular}{>{\centering}p{0.09\textwidth}>{\centering}p{0.09\textwidth}}
  0.89  & 0.98 
\end{tabular}}\\
  & & & 0.8 & \multicolumn{ 3}{l}{\begin{tabular}{>{\centering}p{0.09\textwidth}>{\centering}p{0.09\textwidth}}
  0.56  & 0.62 
\end{tabular}}\\
   
\bottomrule
\toprule
strategy & modality & change class & object type&  \multicolumn{ 3}{c}{\(\text{AP}\)}\\ 
\midrule

\multirow{ 2}{*}{(g)} &\multirow{ 2}{*}{SF} & \multirow{2}{*}{All Ped.-crossing Changes} & lanes & \multicolumn{ 3}{c}{--}   \\
 & & & ped.cross. & \multicolumn{ 3}{c}{0.19}   \\  
   \midrule

\multirow{ 4}{*}{(h)} & \multirow{ 4}{*}{SF} & \multirow{2}{*}{Ped.-crossing Insertions} & lanes & \multicolumn{ 3}{c}{--}   \\
 & & & ped.cross. & \multicolumn{ 3}{c}{0.35}   \\ 
   \cmidrule(){3-7}
 & & \multirow{2}{*}{Ped.-crossing Deletion} & lanes &  \multicolumn{ 3}{c}{--}   \\
 & & & ped.cross &  \multicolumn{ 3}{c}{0.09}   \\
   \midrule

\multirow{ 2}{*}{(i)} &\multirow{ 2}{*}{SF} & \multirow{2}{*}{--} & lanes & \multicolumn{ 3}{c}{0.90}   \\
 & & & ped.cross. & \multicolumn{ 3}{c}{0.82}   \\

\bottomrule
\end{tabular}
\end{table}
\begin{figure}[]
\centering
\begin{subfigure}{\textwidth}
\centering
    \includegraphics[width=\linewidth]{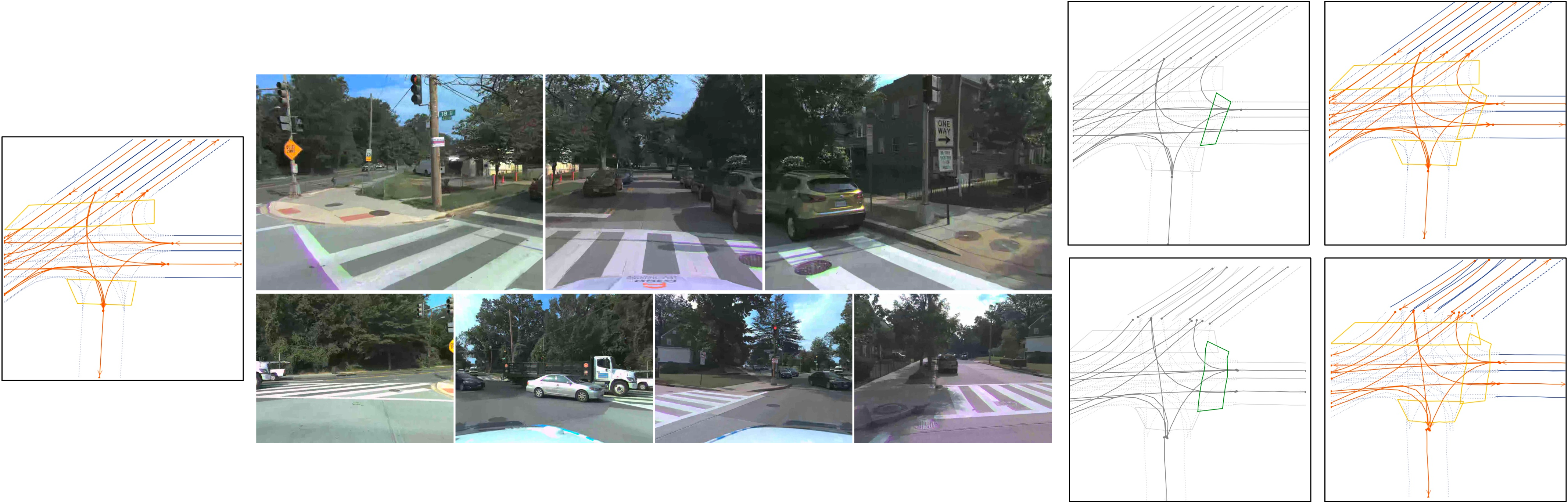}
    \caption{Successful verification + insertion detection}
    \label{fig:inclu1}
\end{subfigure}%

\begin{subfigure}{\textwidth}
\centering
    \includegraphics[width=\linewidth]{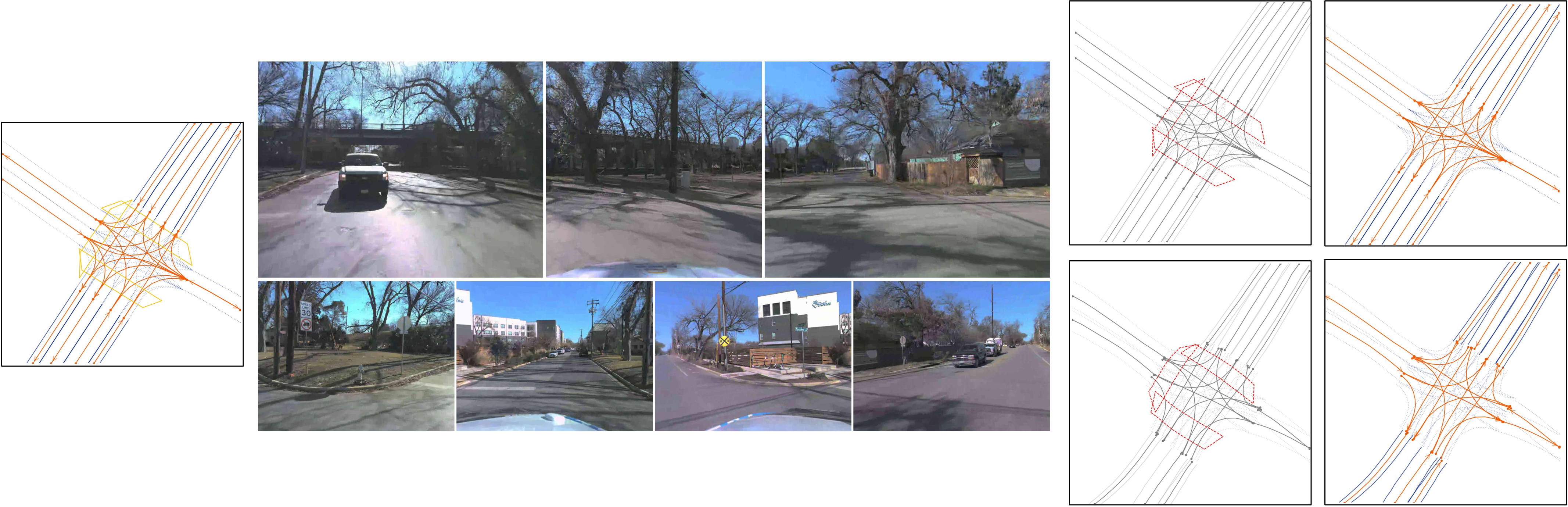}
    \caption{Successful deletion detection}
    \label{fig:inclu2}
\end{subfigure}%

\begin{subfigure}{\textwidth}
\centering
    \includegraphics[width=\linewidth]{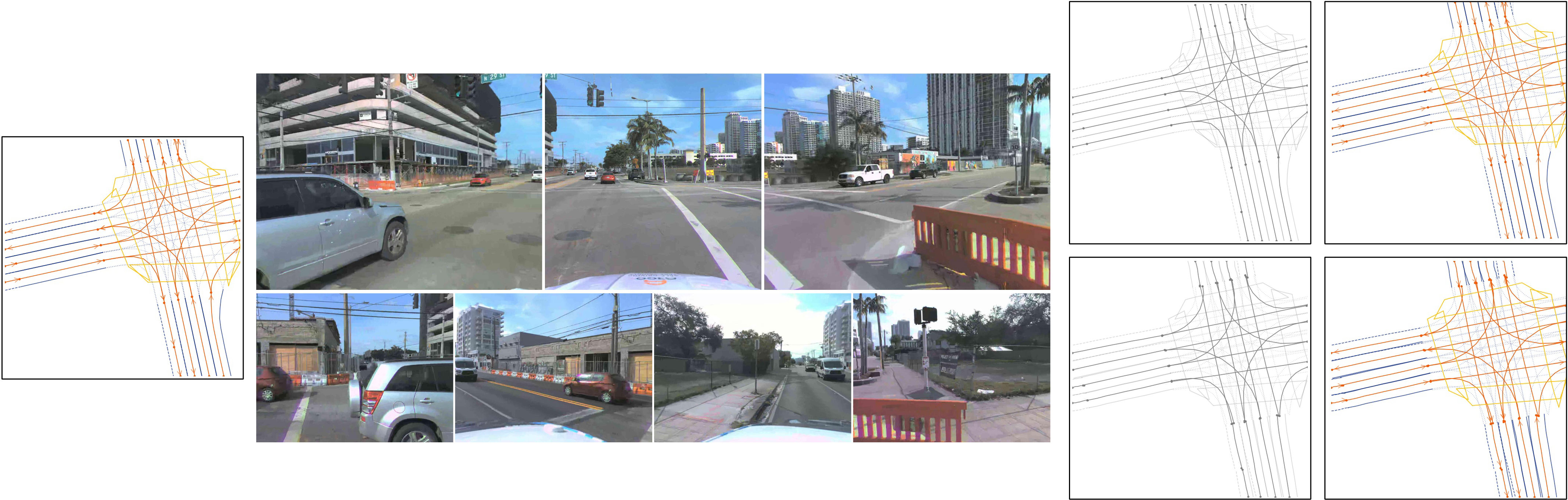}
    \caption{Successful verification}
    \label{fig:inclu3}
\end{subfigure}%

\begin{subfigure}{\textwidth}
\centering
    \includegraphics[width=\linewidth]{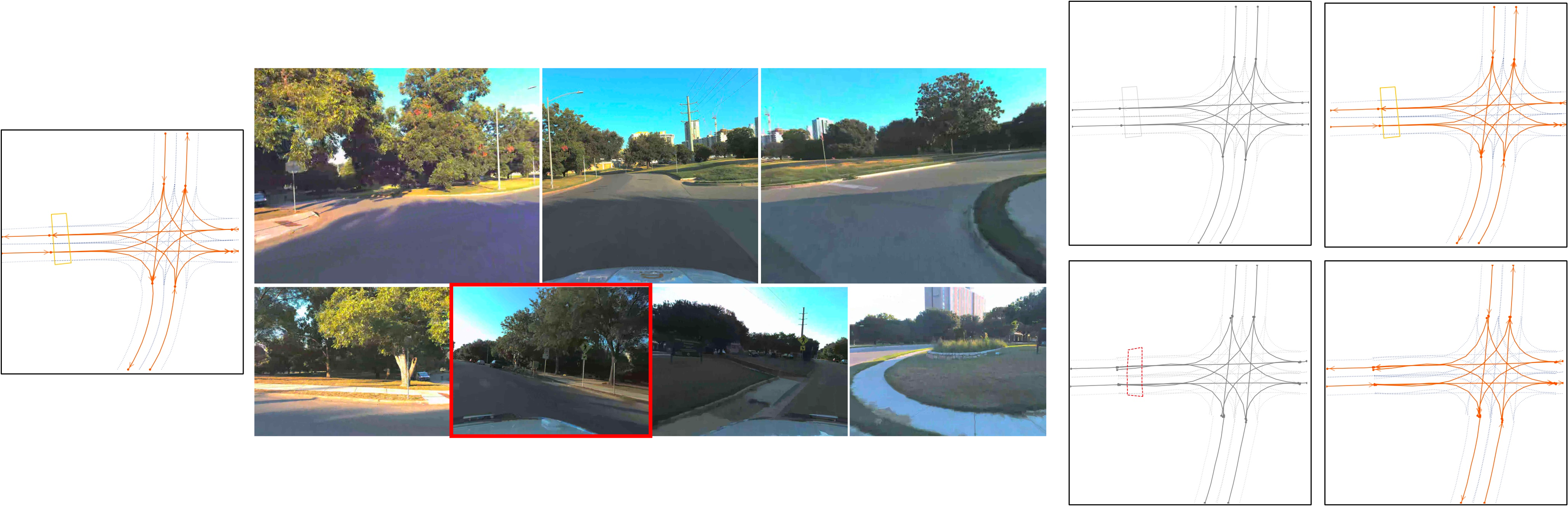}
    \caption{Failure case: Faded pedestrian crossing (see red image)}
    \label{fig:inclu4}
\end{subfigure}%
  \caption{Examples of element-based change detection and update on selected scenes of the Argoverse 2 Map Change Detection Dataset \cite{Argoverse2}. The figures show (from left to right): HD~map prior, camera images, change map (GT and ExelMap) and updated map (GT and ExelMap). In the change map, insertions are color-coded green, deletions dashed-red and unchanged elements grey.
  }
  \label{fig:qualitative}
\end{figure}
\section{Discussion}
In~\cref{fig:qualitative}, examples of element-based change detection related to pedestrian crossings are shown. From left to right, the sub-figures show the stale HD~map, the SF camera input, ground truth and prediction of change scores, and of the local updated map. The change scores are color-coded, with green indicating a pedestrian crossing insertion, dashed-red a deletion and grey no change. We observe that for all lane segments not representing pedestrian crossings, the network successfully learned a pass-through function by cross-attending to the encoded map prior features, although some degradations in lane accuracy can be seen. \par
When looking at the type-aware evaluation strategies in~\cref{table2}, we observe a reduced performance for detecting deleted pedestrian crossings compared to newly inserted ones. This is a consequence of our synthetic dataset being imbalanced towards pedestrian crossing insertions over deletions with a ratio of approximately 9:1. The reason is that insertions can be easily generated by statistically deleting elements from ground truth to generate a synthetic stale map. The complementary procedure cannot be automated and requires manual annotation, leading to the imbalance in the present experiments. Reducing the number of insertions to level the ratio consistently leads to a comparable drop in insertion detection accuracy, which demonstrates the importance of adequate synthetic training data.\par
Although the network is able to capture a large part of pedestrian crossing-related changes, with promising results for element-based change detection, we observe some failure cases. Not surprisingly, changed elements (partially) occluded by other traffic participants or objects are a main source of missed detections for our SF approach. Another consequence of the SF nature of our method are low scores on \(\text{Acc}^{-}\) in MF evaluation. This is expected, as we evaluate only a pseudo-MF version of our network, by summing over all frames \(\mathcal{F}_j\) in the \(i\)-th sequence. When an element is appearing at the edge of the field of view, it is often prematurely classified as changed, which is then corrected in subsequent frames when approaching it. Similarly, actual changes are detected when they appear in the rear or front camera, yet missed when intermediately only appearing partly in the lateral views. This indicates that extending the method to a true MF setting is crucial.\par
Regarding change localization accuracy, we observe a significant number of predicted changes that cannot be attributed to ground truth changes. This is in line with the discussions on premature insertions, with numbers expected to rise in a MF setting. If we compare results for \(\text{Acc}_\text{loca}\) and \(\text{Acc}_\text{loca}^c\) for deleted pedestrian crossings, we see higher numbers for \(\text{Acc}_\text{loca}^c\). This indicates that if we are able to detect deletions, we localize them very well.\par 
Finally, the AP on insertions is comparable to the LaneSegNet backbone, which is consistent, given that these elements are generated purely from sensor data. We observe a slight degradation in accuracy for elements that contain no change, but this could be easily solved in inference by outputting the corresponding element from the stale map, if no change is signalled.

\section{Conclusions and Future Work}
Our approach successfully demonstrates the importance of  element-based change detection for the safety-critical task of HD~map maintenance. We assure explainability through both our network design and detailed evaluation strategies. Although our method is able to detect changes in both lanes and pedestrian crossings, experiments were performed with pedestrian crossing-related changes only due to availability of synthetic training data. Hence, extending the training to more complex change scenarios in the future is crucial, where minor adaptations might be needed in the evaluation strategy. Further advancements may come from refining the encoding and incorporation strategy for the stale HD~map prior, development of a multi-frame formulation of the network and extension to changes in forward view.
\newpage
\section*{Acknowledgements}
The research work was funded by the Swedish Foundation for Strategic Research (SSF) under the project DeltaMap (ID22-0045). This work was partially supported by the Wallenberg AI, Autonomous
Systems and Software Program (WASP) funded by the Knut and Alice
Wallenberg Foundation. This work was financed in part by the Swedish Research Council grant XPLORE3D.

\bibliographystyle{splncs04}
\bibliography{main}

\begin{thebibliography}{10}
\providecommand{\url}[1]{\texttt{#1}}
\providecommand{\urlprefix}{URL }
\providecommand{\doi}[1]{https://doi.org/#1}

\bibitem{bateman2024exploringrealworldmap}
Bateman, S.M., Xu, N., Zhao, H.C., Ben~Shalom, Y., Gong, V., Long, G., Maddern, W.: {Exploring Real World Map Change Generalization of Prior-Informed HD Map Prediction Models}. In: Proceedings of the IEEE/CVF Conference on Computer Vision and Pattern Recognition (CVPR) Workshops. pp. 4568--4578 (2024)

\bibitem{pedcoss}
Bu, T., Mertz, C., Dolan, J.: {Toward Map Updates with Crosswalk Change Detection Using a Monocular Bus Camera}. In: IEEE Intelligent Vehicles Symposium (IV). pp.~1--8 (2023). \doi{10.1109/IV55152.2023.10186622}

\bibitem{nuscenes}
Caesar, H., Bankiti, V., Lang, A.H., Vora, S., Liong, V.E., Xu, Q., Krishnan, A., Pan, Y., Baldan, G., Beijbom, O.: {NuScenes: A multimodal dataset for autonomous driving}. In: Proceedings of IEEE Conference on Computer Vision and Pattern Recognition (CVPR) (2020)

\bibitem{hdmaps}
Elghazaly, G., Frank, R., Harvey, S., Safko, S.: {High-Definition Maps: Comprehensive Survey, Challenges, and Future Perspectives}. IEEE Open Journal of Intelligent Transportation Systems  \textbf{4},  527--550 (2023). \doi{10.1109/OJITS.2023.3295502}

\bibitem{resnet}
He, K., Zhang, X., Ren, S., Sun, J.: {Deep Residual Learning for Image Recognition}. In: IEEE Conference on Computer Vision and Pattern Recognition (CVPR). pp. 770--778 (2016). \doi{10.1109/CVPR.2016.90}

\bibitem{perpx}
Heo, M., Kim, J., Kim, S.: {HD Map Change Detection with Cross-Domain Deep Metric Learning}. In: IEEE/RSJ International Conference on Intelligent Robots and Systems (IROS). pp. 10218--10224 (2020). \doi{10.1109/IROS45743.2020.9340757}

\bibitem{jiang2024pmapnetfarseeingmapgenerator}
Jiang, Z., Zhu, Z., Li, P., ang Gao, H., Yuan, T., Shi, Y., Zhao, H., Zhao, H.: {P-MapNet: Far-seeing Map Generator Enhanced by both SDMap and HDMap Priors} (2024), \url{https://arxiv.org/abs/2403.10521}

\bibitem{tbv}
Lambert, J., Hays, J.: {Trust, but Verify: Cross-Modality Fusion for HD Map Change Detection}. In: Proceedings of the Neural Information Processing Systems Track on Datasets and Benchmarks (NeurIPS Datasets and Benchmarks) (2021)

\bibitem{li2021hdmapnet}
Li, Q., Wang, Y., Wang, Y., Zhao, H.: {HDMapNet: An Online HD Map Construction and Evaluation Framework}. In: International Conference on Robotics and Automation (ICRA). pp. 4628--4634 (2022). \doi{10.1109/ICRA46639.2022.9812383}

\bibitem{li2023lanesegnet}
Li, T., Jia, P., Wang, B., Chen, L., Jiang, K., Yan, J., Li, H.: {LaneSegNet: Map Learning with Lane Segment Perception for Autonomous Driving}. In: International Conference on Learning Representations (ICLR) (2024)

\bibitem{li2022bevformer}
Li, Z., Wang, W., Li, H., Xie, E., Sima, C., Lu, T., Qiao, Y., Dai, J.: {BEVFormer: Learning Bird’s-Eye-View Representation from Multi-Camera Images via Spatiotemporal Transformers}. Proceedings of European Conference on Computer Vision (ECCV)  (2022)

\bibitem{MapTR}
Liao, B., Chen, S., Wang, X., Cheng, T., Zhang, Q., Liu, W., Huang, C.: {MapTR: Structured Modeling and Learning for Online Vectorized HD Map Construction}. In: International Conference on Learning Representations (ICRL) (2023)

\bibitem{maptrv2}
Liao, B., Chen, S., Zhang, Y., Jiang, B., Zhang, Q., Liu, W., Huang, C., Wang, X.: {MapTRv2: An End-to-End Framework for Online Vectorized HD Map Construction}. arXiv preprint arXiv:2308.05736  (2023)

\bibitem{loca}
Lilja, A., Fu, J., Stenborg, E., Hammarstrand, L.: {Localization Is All You Evaluate: Data Leakage in Online Mapping Datasets and How to Fix It}. In: Proceedings of the IEEE/CVF Conference on Computer Vision and Pattern Recognition (CVPR). pp. 22150--22159 (June 2024)

\bibitem{focalloss}
Lin, T.Y., Goyal, P., Girshick, R., He, K., Dollár, P.: {Focal Loss for Dense Object Detection}. IEEE Transactions on Pattern Analysis and Machine Intelligence  \textbf{42}(2),  318--327 (2020). \doi{10.1109/TPAMI.2018.2858826}

\bibitem{vectormapnet}
Liu, Y., Yuantian, Y., Wang, Y., Wang, Y., Zhao, H.: {VectorMapNet: End-to-end Vectorized HD Map Learning}. In: International Conference on Machine Learning. (ICML) (2023)

\bibitem{liu2022bevfusion}
Liu, Z., Tang, H., Amini, A., Yang, X., Mao, H., Rus, D., Han, S.: {BEVFusion: Multi-Task Multi-Sensor Fusion with Unified Bird's-Eye View Representation}. In: IEEE International Conference on Robotics and Automation (ICRA) (2023)

\bibitem{smerf}
Luo, K.Z., Weng, X., Wang, Y., Wu, S., Li, J., Weinberger, K.Q., Wang, Y., Pavone, M.: {Augmenting Lane Perception and Topology Understanding with Standard Definition Navigation Maps}. In: IEEE International Conference on Robotics and Automation (ICRA). pp. 4029--4035 (2024). \doi{10.1109/ICRA57147.2024.10610276}

\bibitem{pannen}
Pannen, D., Liebner, M., Hempel, W., Burgard, W.: How to keep hd maps for automated driving up to date. In: IEEE International Conference on Robotics and Automation (ICRA). pp. 2288--2294 (2020). \doi{10.1109/ICRA40945.2020.9197419}

\bibitem{plachetka}
Plachetka, C., Maier, N., Fricke, J., Termöhlen, J.A., Fingscheidt, T.: {Terminology and Analysis of Map Deviations in Urban Domains: Towards Dependability for HD Maps in Automated Vehicles}. In: IEEE Intelligent Vehicles Symposium (IV). pp. 63--70 (2020). \doi{10.1109/IV47402.2020.9304580}

\bibitem{mindthemap}
Sun, R., Yang, L., Lingrand, D., Precioso, F.: {Mind the map! Accounting for existing map information when estimating online HDMaps from sensor} (2024), \url{https://arxiv.org/abs/2311.10517}

\bibitem{wang2023openlanev2}
Wang, H., Li, T., Li, Y., Chen, L., Sima, C., Liu, Z., Wang, B., Jia, P., Wang, Y., Jiang, S., Wen, F., Xu, H., Luo, P., Yan, J., Zhang, W., Li, H.: {OpenLane-V2: A Topology Reasoning Benchmark for Unified 3D HD Mapping}. In: Proceedings of the Neural Information Processing Systems Track on Datasets and Benchmarks (NeurIPS Datasets and Benchmarks) (2023)

\bibitem{Argoverse2}
Wilson, B., Qi, W., Agarwal, T., Lambert, J., Singh, J., Khandelwal, S., Pan, B., Kumar, R., Hartnett, A., Pontes, J.K., Ramanan, D., Carr, P., Hays, J.: {Argoverse 2: Next Generation Datasets for Self-Driving Perception and Forecasting}. In: Proceedings of the Neural Information Processing Systems Track on Datasets and Benchmarks (NeurIPS Datasets and Benchmarks) (2021)

\bibitem{Yang2022BEVFormerVA}
Yang, C., Chen, Y., Tian, H., Tao, C., Zhu, X., Zhang, Z., Huang, G., Li, H., Qiao, Y., Lu, L., Zhou, J., Dai, J.: {BEVFormer v2: Adapting Modern Image Backbones to Bird's-Eye-View Recognition via Perspective Supervision}. In: IEEE/CVF Conference on Computer Vision and Pattern Recognition (CVPR). pp. 17830--17839 (2023). \doi{10.1109/CVPR52729.2023.01710}

\bibitem{Yuan_2024_streammapnet}
Yuan, T., Liu, Y., Wang, Y., Wang, Y., Zhao, H.: {StreamMapNet: Streaming Mapping Network for Vectorized Online HD Map Construction}. In: Proceedings of the IEEE/CVF Winter Conference on Applications of Computer Vision (WACV). pp. 7356--7365 (2024)

\bibitem{zhou2022crossviewtransformersrealtimemapview}
Zhou, B., Krähenbühl, P.: {Cross-view Transformers for real-time Map-view Semantic Segmentation} (2022), \url{https://arxiv.org/abs/2205.02833}

\bibitem{roundabout}
Zinoune, C., Bonnifait, P., Iba{\~n}ez-Guzm{\'a}n, J.: {Detection of Missing Roundabouts in Maps for Driving Assistance Systems}. In: {IEEE Intelligent Vehicles Symposium (IV)}. pp. 123--128 (2012). \doi{10.1109/IVS.2012.6232245}, \url{https://hal.science/hal-00709919}

\end{thebibliography}
\end{document}